\documentclass[a4paper,fleqn]{cas-dc}

\usepackage[numbers]{natbib}
\usepackage{color,soul}
\usepackage[normalem]{ulem}
\usepackage{float} 
\usepackage{bm}
\restylefloat{table}
\restylefloat*{figure}
\usepackage{graphics}
\usepackage{caption}
\usepackage{subfig}

\newcommand\eg{e.g., }
\newcommand\ie{\textit{i.e.,\ }}
\newcommand\etc{{etc.\ }}

\def\tsc#1{\csdef{#1}{\textsc{\lowercase{#1}}\xspace}}
\tsc{WGM}
\tsc{QE}
\tsc{EP}
\tsc{PMS}
\tsc{BEC}
\tsc{DE}
\begin{document}

\let\WriteBookmarks\relax
\def\floatpagepagefraction{1}
\def\textpagefraction{.001}
\shorttitle{Bridging the Gap}
\shortauthors{Qraitem et~al.}

\title[mode=alt]{Bridging the Gap: Machine Learning to Resolve Improperly Modeled Dynamics}                      
\tnotemark[1,2]

\tnotetext[1]{We gratefully acknowledge the support of ONR Award No. 14-19-1-2253 and NSF DUE Award 1839686.}

\author[1]{Maan Qraitem}
\cormark[1]
\ead{mqrait20@colby.edu }
\address[1]{Department of Computer Science, Colby College, Waterville, ME 04901, USA}

\author[2]{Dhanushka Kularatne}
\ead{dkul@seas.upenn.edu }
\address[2]{Mechanical Engineering and Applied Mechanics, University of Pennsylvania, Philadelphia, PA 19104, USA}

\author[3]{Eric Forgoston}
\ead{eric.forgoston@montclair.edu }
\address[3]{Department of Applied Mathematics and Statistics, Montclair State University, Montclair, NJ 07043, USA}

\author[2]{M. Ani Hsieh}
\ead{m.hsieh@seas.upenn.edu}

\cortext[cor1]{Corresponding author}

\begin{abstract}
We present a data-driven modeling strategy to overcome improperly modeled dynamics for systems exhibiting complex spatio-temporal behaviors.  We propose a Deep Learning framework to resolve the differences between the true dynamics of the system and the dynamics given by a model of the system that is either inaccurately or inadequately described.  Our machine learning strategy leverages data generated from the improper system model and observational data from the actual system to create a neural network to model the dynamics of the actual system.  We evaluate the proposed framework using numerical solutions obtained from three increasingly complex dynamical systems.  Our results show that our system is capable of learning a data-driven model that provides accurate estimates of the system states both in previously unobserved regions as well as for future states.  Our results show the power of state-of-the-art machine learning frameworks in estimating an accurate prior of the system's true dynamics that can be used for prediction up to a finite horizon.
\end{abstract}

% \begin{graphicalabstract}
% \includegraphics{figs/grabs.pdf}
% \end{graphicalabstract}

% \begin{highlights}
% \item Research highlights item 1
% \item Research highlights item 2
% \item Research highlights item 3
% \end{highlights}


\maketitle

% 1. Intro + Lit Review (Ani): Will include motivation and literature review.  We will decide whether this ends up as 1 or 2 sections towards the end.
% 2. Problem Formulation (Dhanushka)
% 3. Methodology (Dhanushka): This section should include any background/tutorial type information on LSTMs or any other relevant ML concepts, algorithms, architecture as well as our proposed network architecture and explanation for why we designed the architecture the way we did.  We will also include the description of the loss functions here.
% 4. Experiments (Ani): This section will include detailed description of the 1D heat equation, Lid cavity equation, and flow over an oscillating cylinder.  This section will also include how we generated the training and testing data, how we train and test, and the metrics we use to evaluate the performance of the network.
% 5. Results (Maan): Reporting of results.
% 6. Discussion (Eric)
% 7. Conclusion and Future Work (Eric)
 
\section{Introduction}

Recent breakthroughs in machine learning (ML) and artificial intelligence (AI) have shown a remarkable ability to extract relationships and correlations in data and events. Indeed, there now exist highly scalable solutions for object detection and recognition, machine translation, text-to-speech conversion, recommender systems, and information retrieval.  Recent advances in machine learning and data analytics have yielded transformative results across diverse scientific disciplines \cite{ref:Alipanahi2015,ref:Krizhevsky2012,ref:Lake2015,lecun2015deep,ghahramani2015probabilistic}. Enabled by the decreasing price to performance ratio of sensing, data storage, and computational resources in the past decade, data-driven machine learning strategies are taking center stage across many scientific disciplines. 

In the realm of complex spatiotemporal dynamical systems, data-driven machine learning strategies have been employed for reduced-order models (ROMs) \cite{Bhatnagar2019,Wiewel2019,white2019fast,mohan2018deep,lee2017prediction}, discovery of system dynamics \cite{Tibshirani1996,YAO2007,Bongard2007,Schmidt2009,Kim2016,Pan16,Brunton16,lusch2017deep,maulik2018data,maulik2019subgrid,ayed2019learning,almomani2020entropic}, computation of dynamical system solutions \cite{raissi2017PINN1,raissi2017PINN2,raissi2018deep,raissi2018hidden,raissi2018multistep,pathak2018hybrid}, and prediction of future dynamics \cite{Ling2015,miyanawala2017efficient,raissi2018hidden,pathak2018hybrid,viquerat2019,LAPEYRE2019}.  These recent developments spurred by the current enthusiasm surrounding ML and AI strategies can be broadly classified into two categories: works that investigate the feasibility of existing ML/AI algorithms and architectures, and those centered around the development of new algorithms and architectures.  Existing work whose main objective is the former have focused on the power of ML/AI techniques to significantly reduce the steep computation and data storage costs associated with high-fidelity computational fluid dynamics (CFD) efforts \cite{Bhatnagar2019,Wiewel2019,white2019fast,mohan2018deep,lee2017prediction,Ling2015,miyanawala2017efficient,raissi2018hidden,viquerat2019,LAPEYRE2019}.  These works often leverage existing CFD models to generate ground truth, training, and testing data sets to evaluate well-studied convolutional neural networks (CNN) \cite{Bhatnagar2019,LAPEYRE2019,miyanawala2017efficient}, long short-term memory (LSTM) networks \cite{mohan2018deep}, generative adversarial networks (GAN) \cite{lee2017prediction}, and existing ML/AI frameworks \cite{Ling2015,white2019fast,viquerat2019}.  Nevertheless,  existing ML/AI strategies are predicated on access to large amounts of {\it labeled} data where explicit knowledge derived from well-established first principles are difficult to encode. 

Works in the second category that directly address these challenges include sparse regression techniques \cite{YAO2007,Kim2016,Pan16,Brunton16,lusch2017deep,almomani2020entropic} and physics-informed neural networks (PINNs) \cite{raissi2017PINN1,raissi2017PINN2,raissi2018deep,raissi2018hidden}.  Sparse identification is a data-driven system identification strategy that balances model complexity with descriptiveness \cite{Brunton16}. Since the dynamics of most physical systems are governed by only a few important terms \cite{Brunton16}, sparse identification selects from a finite set of candidate dictionary functions whose linear combination describes the system dynamics \cite{Pan16}.  On the other hand, PINNs are neural networks that are trained to solve supervised learning tasks whose dynamics can be described by general nonlinear PDEs. The key advantage of PINNs is their {\it data-efficiency} in the training phase. Sparse regression techniques such as those found in \cite{YAO2007,Kim2016,Pan16,Brunton16,lusch2017deep,almomani2020entropic} require large amounts of relatively clean data to accurately compute numerical gradients, whereas PINNs do not require any data on gradients of the flow field (nor their numerical approximations).  As such, PINNs perform more robustly when data is sparse and/or noisy relative to the complexity of the underlying system dynamics \cite{raissi2017PINN1,raissi2017PINN2}.  In contrast, \citeauthor{ayed2019learning} uses actual observations of a system whose dynamics are given by an ordinary differential equation to train the neural network weights.  Once trained, the network provides an equation-free model representation of the system dynamics.  Different from \cite{YAO2007,Kim2016,Pan16,Brunton16,lusch2017deep,raissi2017PINN1,raissi2017PINN2,almomani2020entropic}, the work does not directly address the issue of data-efficiency but assumes the network has access to a sufficiently large set of training data.

In this work, we take inspiration from \cite{YAO2007,Kim2016,Brunton16,raissi2017PINN1,raissi2017PINN2,maulik2018data,pathak2018hybrid,maulik2019subgrid,almomani2020entropic} and present a data-driven Deep Learning framework capable of resolving the differences between the actual dynamics of a complex nonlinear system and that of the same system which has been improperly or inaccurately modeled.  Given an inaccurate or inadequate model of a system,  our proposed ML strategy combines data from this inaccurate/inadequate model with observational data from the actual system to learn the dynamics of the actual system.  The result is a neural network model that can accurately estimate the system states in regions with no observations and/or provide predictions for future states.  Different from \cite{YAO2007,Kim2016,Brunton16,raissi2017PINN1,raissi2017PINN2,almomani2020entropic}, our approach provides an equation-free representation of the system dynamics  that successfully estimates the underlying physics that drives the process. We evaluate the proposed framework using three different dynamical systems each with increasing complexity.  Our results show how the proposed strategy is not only capable of resolving improperly or inaccurately modeled dynamics but also can learn the dynamics of the actual system and provide accurate future predictions.  

While our approach is similar to \cite{pgnn_lake,pathak2018hybrid}, we make use of LSTMs in our deep learning network rather than a simple multi-layer perceptron \cite{pgnn_lake} or reservoir computer \cite{pathak2018hybrid}. Our approach is general and may be used for a wide range of dynamical systems of different dimension and complexity, including examples in which the known model is missing external forcing functions or other known dynamics. Even for these complicated scenarios, we demonstrate in this article the power of our method to successfully predict the dynamics wherein simpler approaches will fail. Since our output is a neural network representation of the system model, the output of our network  can be fed into existing data-driven model discovery techniques \cite{Tibshirani1996,Bongard2007,Schmidt2009,Brunton16,Pan16} to obtain closed-form equation representations of the dynamical system. 

%This work will leverage these capabilities %to identify correlations and extract new physical insights that are not captured by existing models of the LCS.  The idea is 
%to develop a data-driven framework to ...%identify the relative importance of the various interdependent physical processes, \eg inflow/outflow and bathymetry, that drives LCS dynamics and be incorporated into the physical expressions of existing models (Recommendations 23, 25 in \cite{ref_NAP}).  
%Nevertheless, ML/AI strategies are heavily reliant on large repositories of {\it labeled} data and are often computationally expensive.  Therefore, our proposed strategy is to focus on techniques that fuse the generalization capabilities of neural network based machine learning techniques with first principles based modeling of the physical phenomena to improve forecasting capabilities.  

%\section{Literature}

%As such, the ability to encode prior knowledge, physical constraints, and domain expertise is paramount for any successful application of machine learning strategies to the study GFD flows.  

The paper is organized as follows: we list our assumptions and provide a concise formulation of our problem in Sec. \ref{sec_problem}.  The design of the network architecture and our methodology is described in Sec. \ref{sec_architecture}.  We discuss how we evaluate our methodology in Sec. \ref{sec_eval} and present our results with discussion in Sec. \ref{sec_results}.  Conclusions and directions for future work are contained in Sec. \ref{sec_conclusion}.

\section{Problem Formulation}\label{sec_problem}
We consider a spatio-temporal process $u(x,t) \in \mathbb{R}^m$, where $x\in \mathbb{W}$ represents a point in the environment $\mathbb{W} \subset \mathbb{R}^n$ and $t\in [t_s,\; t_f]$ represents the time within an observation interval of interest.  The actual model of the process that governs $u$ is denoted by $M_{act}$ and is given by a partial differential equation (PDE) of the form
\begin{equation}
    \label{eqn_sec_prob_act_model}
    u_t =\mathcal{N}[u,f_1,\cdots,f_p,g_1,\cdots,g_r], 
\end{equation}
where $\mathcal{N}[\cdot]$ is a nonlinear differential operator, where $f_i = f_i(x,t) \in \mathbb{R}^{n_{f_i}},\; i=1,\ldots,p$ and $g_i = g_i(x,t) \in \mathbb{R}^{n_{g_i}},\; i=1,\ldots, r$ are external phenomena that impact $u$.  Let $M_{curr}$ denote the model that is obtained from the current understanding of the physics of $u$. Then $M_{curr}$ is given by the PDE with form
\begin{equation}
    \label{eqn_sec_prob_curr_model}
    u_t = \tilde{\mathcal{N}}[u,f_1,\cdots,f_p],
\end{equation}
where $\tilde{{\cal N}}[\cdot]$ is also a nonlinear differential operator.  Here, the $f_i$ denote the $p$ external phenomena whose impact on $u$ are currently {\it known} and the $g_i$ denote the $r$ external phenomena that affect $u$ but are not captured in $M_{curr}$.  Note that in general, $g_i$ could represent some error in $f_i$ so that $g_i = f_i + \epsilon$ where $\epsilon$ denotes the difference between $f_i$ and $g_i$.  Furthermore, $\tilde{{\cal N}}$ is used to denote any differences in system parameters between $M_{curr}$ and $M_{act}$.  Thus, while $M_{curr}$ represents the current understanding of the process, this understanding is incomplete or inadequate and thus $M_{curr}$ is not an accurate representation of the process model.   

%We assume that a model $M_{curr}$, obtained from the current understanding of the physics of $u$ exists, and that $M_{curr}$ is expressed in the form of a partial differential equation (PDE), given by
%\begin{equation}
%    \label{eqn_sec_prob_curr_model}
%    u_t = \tilde{\mathcal{N}}[u,f_1,\cdots,f_p]
%\end{equation}
%where $\tilde{\mathcal{N}}[\cdot]$ is some nonlinear (differential) operator and $f_i = f_i(x,t) \in \mathbb{R}^{n_{f_i}},\; i=1,\ldots,p$ are $p$ external phenomena who's impact on $u$ are currently known.

%While $M_{curr}$ represents the current understanding of the process, we assume that this understanding is incomplete, and that the actual model $M_{act}$ that governs $u$ is given by the PDE,
%\begin{equation}
%    \label{eqn_sec_prob_act_model}
%    u_t =\mathcal{N}[u,f_1,\cdots,f_p,g_1,\cdots,g_r], 
%\end{equation}
%where $\mathcal{N}[\cdot]$ is also a nonlinear (differential) operator, and $g_i = g_i(x,t) \in \mathbb{R}^{n_{g_i}},\; i=1,\cdots r$ are $r$ external phenomena that affect $u$ but are not captured in $M_{curr}$. Note that $g_i$ could represent some error in $f_i$,  so that $g_i = f_i + \epsilon$ where $\epsilon$ denotes the error.

Given a set of coordinates $S = \{s_j\vert s_j = (x_j, t_j),\; x_j\in \mathbb{W},\; t_j \in [t_s,\;t_f],\;j=1,\ldots, n_{data}\}$, let $\hat{U}_{act} = \{\hat{u}_{act_j}\vert j=1,\ldots, n_{data}\}$ be the set of observations of $u$ obtained by measuring the actual process at coordinates $s_j \in S$.  Similarly, let $U_{curr} = \{u_{curr_j}\}$ and $U_{act} = \{u_{act_j}\}$ be the solution sets obtained from $M_{curr}$ and $M_{act}$ respectively, at the coordinates in $S$. In this work, $U_{act}$ is based on computer simulations, but could in fact be measured experimentally. For simplicity, we assume that there are no measurement errors, \ie $\hat{u}_{act_j} \equiv u_{act_j}$ for each $\hat{u}_{act_j} \in \hat{U}_{act}$ and  $u_{act_j} \in U_{act}$ obtained at the same coordinate $s_j \in S$. 

Given $U_{act}$, $U_{curr}$ and observations of a subset of the $g_i$ at the coordinates in $S$, the objective of this work is to develop a neural network based model $M_{nn}$ that better estimates the process $u$ in and potentially beyond the space-time domain $\mathbb{W}\times [t_s,\;t_f]$. Let $e_* = \Vert M_{act} -M_* \Vert \geq 0$ represent some measure of the error of the output of a given model with respect to the output of $M_{act}$ in a given domain. We want $e_{nn} \leq e_{curr}$ in all domains (ideally $e_{nn} = e_{curr}$ only when $e_{curr}=0$), \ie the neural network should be much better at predicting/estimating $u$ than the existing model. %The error measures used for evaluation of the models are described in section \ref{subsec_eval}.

%\ani{@Maan and Dhanushka: $M_{curr}$, $M_{act}$, and $M_{nn}$ are not referred to in Methodology section.  This needs to be fixed.  There is a confusion between what each of these are and need to better defined.}
%Let the set of solutions obtained from this model at the coordinates in $S$ be $U_{nn} = \{u_{nn_j}\}$. Denoting the predictive errors of $M_{curr}$ and  $M_{nn}$ as $e_{curr_j} = (u_{data_j} - u_{curr_j})$ and $e_{nn_j} = (u_{data_j} - u_{nn_j})$ respectively for $j=1,\cdots, n_{data}$, we want $\sum_{j=1}^{n_{data}} \|e_{nn_j}\| \ll \sum_{j=1}^{n_{data}} \|e_{curr_j} \|$, \ie the neural network should be much better at predicting $u$ than the existing model. 

To illustrate, consider a mass-spring-damper system with mass $m$, damping coefficient $c$ and spring constant $k$ that is subjected to two external forcing functions given by $F_1(t) = A_1 \cos{(\omega_1 t)}$ and $F_2(t) = A_2 \cos{(\omega_2 t)}$. If the displacement of the mass is denoted by $y$, the actual model of the system $M_{act}$ is given by the ordinary differential equation (ODE)
\begin{equation}
    \label{eqn_prob_msd_approx}
    m\Ddot{y} + c\dot{y} + ky = F_1 + F_2.
\end{equation}
Let's assume that due to modeling and measurement errors, the model that we have access to, $M_{curr}$, is given by
\begin{equation}
    \label{eqn_prob_msd_curr}
    \tilde{m}\Ddot{y} + \tilde{c}\dot{y} + \tilde{k}y = F_1.
\end{equation}
Note that this model only captures part of the forcing function and has errors in the mass, spring, and damping coefficients. Given measurements of the displacement $y$, our work seeks to develop a neural network, whose output closely resembles that of the actual model $M_{act}$ for the same initial conditions. Denoting the output of the actual, current and neural network models by $y_{act}(t)$, $y_{curr}(t)$ and $y_{nn}(t)$ respectively, we would like $\Vert y_{nn}(t) - y_{act}(t)\Vert < \epsilon < \Vert y_{curr}(t) - y_{act}(t)\Vert$, where ideally $\epsilon$ is small. In other words, we would like the trained neural network output to always be a better approximation of the ground truth than the current model output or match the ground truth exactly.  Lastly, in our proposed framework, the neural network model $M_{nn}$ only provides outputs for the ODE, {\it e.g.}, $y$, $\dot{y}$, and $\ddot{y}$ rather than the equation of the actual ODE.
%\ani{@Eric and Dhanushka: I think this is what Dhanushka meant originally.  Does this sound right? Dhanushka: yes this is a way better way of saying it :-)}

%That is, at the end of the training process, the neural network output \sout{should ideally}\changeto{would} be a solution to an ODE of the form 
%\begin{equation}
%    \label{eqn_prob_msd_nn}
%    (m+\delta_m)\Ddot{y} + (c+\delta_c)\dot{y} + (k+\delta_k)y = F_1 + F_2 + %\delta_{F_2}.
%\end{equation}
%\changeto{where $\delta_m$, $\delta_c$ and $\delta_k$ are model parameter error terms, and $\delta_{F_2}$ is the modeling error for the second forcing term. Ideally, at the end of the training phase, the magnitude of these error terms should be small}. \sout{We denote $M_{nn}$ to be the differential model represented by the ODE in \eqref{eqn_prob_msd_nn}}\changeto{In the context of this example, $M_{nn}$ represents ODE of the trained model, i.e. the ODE in \eqref{eqn_prob_msd_nn}}. It should be noted that the actual trained network will only provide the solutions to this ODE, and not the ODE itself. Thus, $M_{nn}$ is abstractly referred to as the model that the neural network implements. \ani{I think we need to be careful here because we do not know whether $M_{nn}$ is necessarily outputting data from a model explicitly of the form given by equation \ref{eqn_prob_msd_nn}.  I think it is more correct to say that $M_{nn}$ would be outputting data that {\it approximates} a system of the form given by equation \ref{eqn_prob_msd_nn}.  Is this correct or splitting hairs?}

\section{Methodology}\label{sec_architecture}
%\ani{@Maan and Dhanushka: We should describe the network architecture we are using in this section.  I am thinking about the block diagram that Maan had in one of his PPT presentations.}

The proposed method uses a neural network based framework to ``bridge the gap'' between $M_{curr}$ and $M_{act}$. Neural networks have recently been used in a plethora of prediction and estimation problems. However, in most of these solutions, large quantities of training data is required to obtain good prediction performance. This is especially true for prediction/estimation problems involving complex dynamical systems. In this work, we mitigate this data inefficiency problem by incorporating existing knowledge of the process into the neural network architecture. 

The fundamental hypothesis of our work is that the current understanding of the physics of $u$ given by $M_{curr}$, has substantial information that the neural network can exploit in order to provide better predictions of the process. Thus, in addition to the space-time coordinates ($x,\;t$) and, where applicable, external forcing terms $g_i$, we also use the output from $M_{curr}$ as an input to the neural network. This input may be presented to the network in different formats, \eg data generated from a reduced-order model \cite{KirbyBook2000,Sirovich1987TurbulenceStructures}, coefficients and functions from a sparse identification of the process \cite{Brunton16,almomani2020entropic}, output data from a numerical model, \etc

Furthermore, the behaviour of any dynamical system depends heavily on the initial and boundary conditions. In the absence of explicit initial and boundary conditions, these spatio-temporal dependencies have to be captured by the network in a purely data-driven manner. We facilitate this by 1) using Long Short-Term Memory (LSTM) stages in our network to capture temporal dependencies, and 2) providing the network with data in a space-time hypercube around the point of interest. 

\subsubsection*{Neural Networks and LSTM Networks}
%\ani{@Maan -- Can we add a paragraph or two here on the basics of a neural network?  So for example terminologies such as neurons, layers, connectedness of a network, activation functions, backpropagation, and so on should be provided.  We should also include a standard picture that most intro to NN tutorials provide.  Basically, something that summarizes the section titled ``The Neural Network Model'' on this page \url{http://ufldl.stanford.edu/tutorial/supervised/MultiLayerNeuralNetworks/}.  And end the section saying that these are solved using backpropagation and add a cite.  You can even use the webpage as the citation.}

Artificial neural networks (ANN) are powerful nonlinear statistical models which consist of multiple layers of interconnected nodes such that every connection represents a weight. Each node calculates a weighted sum of the outputs of neurons which are connected to it as well as a bias term. By representing the system in terms of layers, neural networks are able to learn features exhibited by highly nonlinear and complex data in a powerful hierarchical fashion. The nonlinearity of these networks comes from the use of nonlinear activation functions in the neural net nodes. The neural net is trained  by minimizing a loss function. The minimization is commonly done by a gradient-based optimization algorithm that makes use of backpropagation -- a computationally efficient algorithm that computes the gradient of the loss function with respect to the weights at each layer. Common optimization algorithms include stochastic gradient descent, Adam \cite{adamoptim}, and Adagrad \cite{adagradoptim}. The optimization algorithm commonly performs updates to the weights using batches of the dataset. A complete pass through all the dataset batches is usually referred to as an {\it epoch}.

\begin{figure}
    \centering
    \includegraphics[width = 0.7\linewidth]{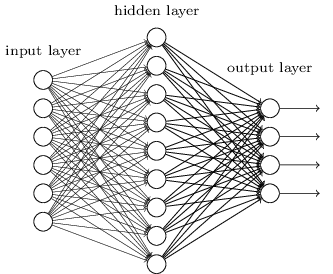}
    \caption{A general dense layer architecture.} 
    \label{FC-arch}
\end{figure}

The most basic structure of a neural network is a fully connected or dense ANN as displayed in Fig. \ref{FC-arch}. Each node in the neural network is governed by an activation function 
%\begin{equation}
    $a_{l+1}(W_l  a_{l} + b_l)$ 
%    \label{eq:nne}
%\end{equation}
where $W_l$ and $b_l$ denotes the weights matrix and bias vector for layer $l$ respectively.  Common choices for $a_{l+1}$ include the sigmoid function commonly denoted by $\sigma(\cdot)$, the hyperbolic tangent function $tanh(\cdot)$, and rectified linear unit function $ReLU(\cdot)$. We refer the reader to \cite{activationfunctions} for a detailed review of activation functions.

In choosing a neural network architecture, we make note that our problem is in nature time-dependant. More concretely, the problem imposes an order on the sequence of observations that must be preserved. In general, standard artificial neural networks are not well-suited to learn such orders since the weights in each ANN layer are fully connected to the previous layer. This forces the ANN to  consider the entire sequence at once.  Recurrent Neural Networks (RNNs), on the other hand, are a different type of neural network that is well suited for sequence learning problems. They are equipped with a memory unit which is updated for each new observation. Thus, parameters of the network are shared for each step in the sequence. As such, RNNs rather than ANNs are most commonly employed to learn time dependencies.

\begin{figure}
    \centering
    \includegraphics[width=0.95\linewidth]{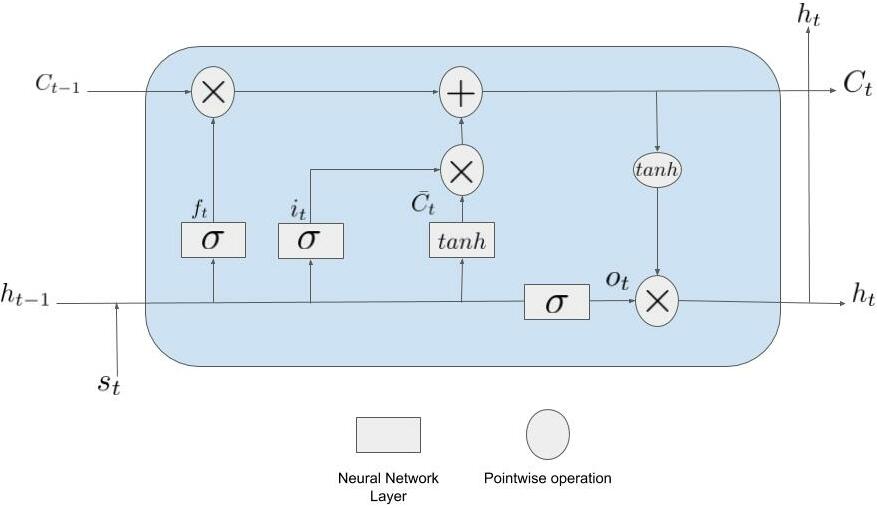}
    \caption{A general architecture for an LSTM layer.} 
    \label{Lstm-arch}
\end{figure}

The Long Short-Term Memory (LSTM) network is a variant of RNNs.  LSTMs address the bottlenecks in traditional RNNs such as the vanishing gradient problem  \cite{vanishinggradient} which hampers learning of long data sequences.  The LSTM memory unit is usually called the cell, denoted by $C$, which is regulated by three gates: an input gate ${\cal I}$, a forget gate ${\cal F}$, and an output gate ${\cal O}$. The input gate controls the contribution of the input to the cell, the forget gate controls what parts of the cell to keep, and the output gate controls the contribution of the cell to the output of the LSTM. A schematic of the architecture can be found in Fig. \ref{Lstm-arch}, with $h$ representing the output of the network while the input of the network is represented with $s$. The equations to compute the gates and states are given by
\begin{align}
    \label{eq:gs}
    {\cal F}_t & = \sigma(W_{\cal F} \cdot [h_{t-1}, s_t] + b_{\cal F}]), \nonumber \\
    {\cal I}_t & = \sigma(W_{\cal I} \cdot [h_{t-1}, s_t] + b_{\cal I}), \nonumber \\
    \bar{C_t} & = tanh(W_C \cdot [h_{t-1} ,s_t] + b_C), \nonumber \\
    C_t & = {\cal F}_t * C_{t-1} + {\cal I}_t*\bar{C_t}, \\ 
    {\cal O}_t & = \sigma(W_{\cal O} \cdot [h_{t-1}, s_t] + b_{\cal O}), \nonumber \\
    h_t & = {\cal O}_t*tanh(C_t), \nonumber
\end{align}
where $\bar{C}$ is the updated state, $W$ is the weights matrix, $b$ is the bias vector for each gate, $s_t$ is the input to the network at time $t$, and * denotes the Hadamard product.  The forget gate reduces overfitting by controlling how an incoming input contributes to the hidden state. This structure is the key reason why LSTMs do not suffer from the vanishing gradient problem exhibited by RNNs. For more detailed discussions on ANNs, RNNs, and LSTMs, we refer the interested reader to \cite{Goodfellow-et-al-2016,lstmexplained,SHERSTINSKY2020132306,chang2018antisymmetricrnn}. 

\subsubsection*{Input data format to the network}
\begin{figure*}
    \centering
    \subfloat[]{\includegraphics[width = 0.4\linewidth]{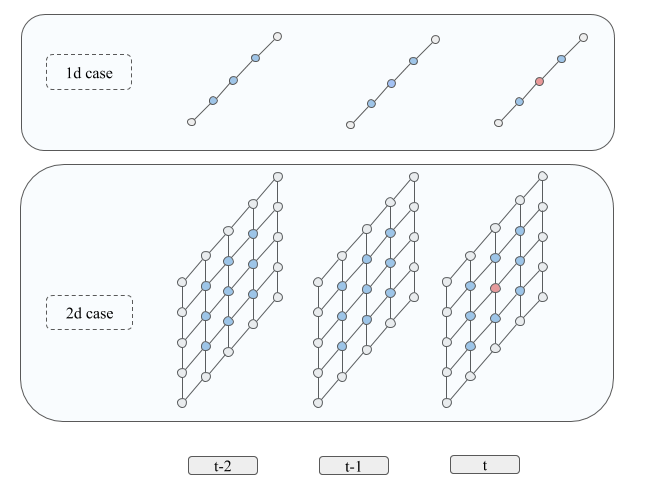}\label{fig_method_data_format}}
    \subfloat[]{\includegraphics[width=0.6\linewidth]{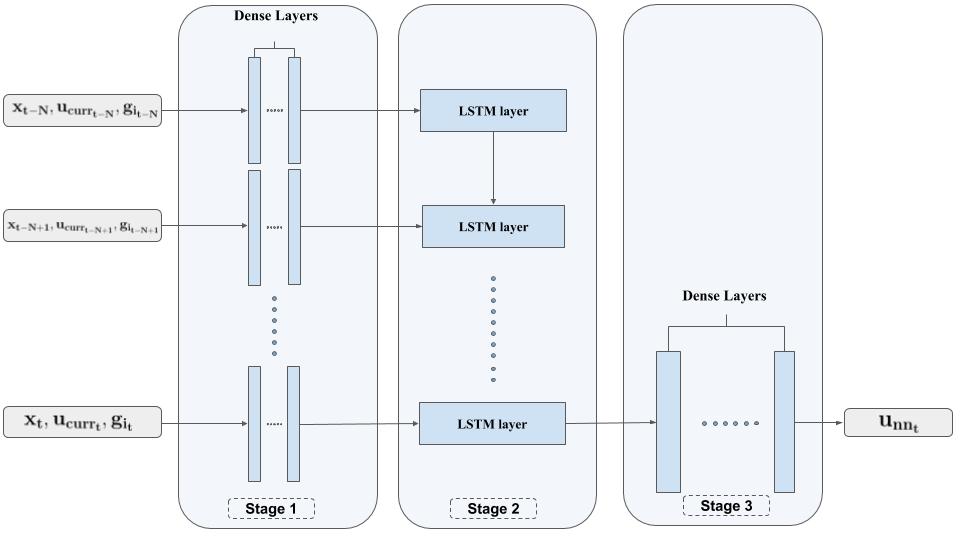}\label{Network-arch}}
    \caption{(a) Format of the inputs to the network. For each input to the network, we consider data in a $n+1$ hypercube around the point of interest. Along each dimension, $k$ data points are included resulting in $k^{n+1}$ number of data points for each input. In this figure, $k=3$.  (b) The general architecture of our final model which is composed of three stages. The number of layers and nodes in each stage depends on the problem.}
\end{figure*}

The network predicts/estimates the process on a point by point basis. In order to capture the spatio-temporal dependencies between the inputs and the output at each coordinate $j$, we consider a $n+1$ dimensional space-time hypercube of the inputs around this coordinate. We consider $k$ data points along each dimension, resulting in  $k^{n+1}$  number of data points for each input.  In general, the larger the choice of $k$, the larger the input data and thus the higher the computational load.  In this work, we choose $k=3$ to limit the computational burden.  Thus for scenarios where $n=2$, as shown in Fig. \ref{fig_method_data_format}, we would consider a hypercube with 27 vertices for each input. 

\subsection{Architecture of the Neural Network}
Our proposed neural network architecture is composed of three stages as shown in Fig. \ref{Network-arch}. We modify the architecture for each problem by changing the number of layers/nodes at different stages of the architecture. The three stages of the network are: \begin{itemize}
    \item Stage 1: Time distributed dense stage with $D_1$ layers;
    \item Stage 2: Long Short-Term Memory (LSTM) stage with $D_2$ layers; and
    \item Stage 3: Dense output stage with $D_3$ layers.
\end{itemize}
The three stages are described below in detail.

\subsubsection*{Stage 1: Time Distributed Dense Layers}
This stage consists of a set of parallel dense layers that work on the inputs at each time slice independently. The purpose of this stage is to give the network the ability to pre-process the data and learn a representation that is most optimal for the LSTM stage. While most research in the literature employing LSTM networks do so without this pre-processing layer, our experiments have demonstrated that adding this stage improves the convergence of the network. The activation function for layer $l$ in this stage is denoted as $a_{l, t}$ where $t$ denotes the time step. In this case, $W$ and $b$ are shared for each time step. The output of this layer is then passed to Stage 2.

\subsubsection*{Stage 2: LSTM Stage}
The LSTM is a type of Deep Learning architecture that is designed to exploit long term dependencies in time series data. Given the nature of dynamical systems data where time-based dependencies are abundant, LSTMs are a powerful choice to model such data. Thus, after the data has been processed by a sequence of dense layers in Stage 1, we apply a sequence of LSTM layers in Stage 2. The equations for the LSTM layer are given by Eq. (\ref{eq:gs}) with $s_t$ replaced by $a_{L, t}$, where $L$ is the number of the last layer in Stage 1.  The output of the LSTM layer from the final time step is then used as the input to the Stage 3.

\subsubsection*{Stage 3: Dense Output Stage}
Stage 3 consists of a sequence of dense layers. This stage serves as a final stop for processing the data before producing the output. The output of the last dense layer is the final predicted output $u_{nn}$ from the neural network. The output of the network is used in the following loss function to train the network
\begin{equation}
    Loss(u_{act}, u_{nn}) = \frac{1}{M} \sum_{i=1}^{M}  (u_{{act}_{i}} - u_{{nn}_{i}})^2,
\end{equation}
where  $M$ is the dimension of the output $u$. 

\section{Methodology Evaluation}\label{sec_eval}
To quantitatively and qualitatively evaluate our methodology, we consider different dynamical systems each with increasing complexity.  The proposed learning framework is evaluated with respect to its ability to reproduce the dynamics of the actual system and its ability to predict future observations on a point-by-point basis.  

\subsection{Candidate Systems}
We consider three candidate systems to test our hypothesis on, with each system being progressively more complex. Each candidate system exhibits one of the three types of differences between $M_{act}$ and $M_{curr}$: 1) differences in system parameters, {\it e.g.}, $u_t = \tilde{\mathcal{N}}[u,f_1,\cdots,f_p]$ with $g_i=0$ for all $i=1, \ldots, r$; 2) differences in external forcing functions and/or boundary conditions, {\it e.g.}, $u_t = \tilde{\mathcal{N}}[u,f_1,\cdots,f_p]$ with $g_i = f_i + \epsilon$ for $i=1, \ldots, r$ with $r \leq p$; and 3) missing terms in the partial differential equation describing the dynamics of the system, {\it e.g.}, $u_t = \tilde{\mathcal{N}}[u,f_1,\cdots,f_p]$ with $g_1\neq 0$.  We briefly describe the candidate systems below. 

\subsubsection*{System 1: 1D Heat Equation} In our first system, we assume both the actual model $M_{act}$ and current model $M_{curr}$ system dynamics are given by
\begin{equation}
    \label{eqn_sec_prob_1D_heat}
    u_t = D_{*}u_{xx},
\end{equation}
%and $M_{act}$ is given by
%\begin{equation}
%    \label{eqn_sec_prob_1D_heat_M_act}
%    u_t = D_{act}u_{xx},
%\end{equation}
where $x \in \mathbb{R}$, $u \in \mathbb{R}$ is the temperature, and $D_{*}$ is the diffusion coefficient and is set to either $D_{act}$ or $D_{curr}$. In this scenario, the discrepancy in the models arise due to a mismatch in the actual and assumed diffusion coefficients. %\changeto{$D_{act} = 15$ while $D_{curr} = 1$}

\subsubsection*{System 2: Lid Cavity Problem}
For our second system, we consider a modified version of the lid cavity problem presented in \cite{2DunsteadyNavierStokes}. The actual model, $M_{act}$, is given by 
\begin{equation}
    \label{eqn_sec_prob_lid_cav_M_act}
    u_t = -(u\cdot\nabla)u -\nabla p +\frac{1}{Re}\nabla^2u + F.
\end{equation}
In \cite{2DunsteadyNavierStokes}, $F$ is chosen to be an external body force with a whirlpool effect. In this work, we employ the same $F$ as in \cite{2DunsteadyNavierStokes} but include a periodic element to $F$ whose components are given by
\begin{align*}
F_x &= (12-24y)x^4+(-24+48y)x^3+\\ 
&(-48y+72y^2-48y^3+12)x^2+ \\
&(-2+24y-72y^2+48y^3)x+\\
&(1-4y+12y^2-8y^3)120\sin{(e^{1.3t} + 80t)},\\
F_y & = (8-48y+48y^2)x^3+ \\
        & (-12+72y-72y^2)x^2+\\
        &(4-24y+48y^2-48y^3+24y^4)x+\\
        &(-12y^2+24y^3-12y^4)120\cos{(e^{1.3t} + 80t)}.
\end{align*}

In this system, the assumed model, $M_{curr}$, is given by the Navier-Stokes equation for incompressible flows,
\begin{equation}
    \label{eqn_sec_prob_lid_cav_M_curr}
    u_t = -(u\cdot\nabla)u -\nabla p +\frac{1}{Re}\nabla^2u
\end{equation}
with $\nabla u = 0$, where $x\in\mathbb{W}\subset\mathbb{R}^2$ denotes the position, $u\in \mathbb{R}^2$ is the flow velocity, $Re$ is the Reynolds number, and $p$ is the pressure. In contrast to the classical lid cavity problem, where the domain $\mathbb{W}$ is a square in which the top boundary moves with a constant speed, we assume the dynamics are subject to periodic boundary conditions at the top and bottom boundaries of the square given by 
\begin{align*} 
    u_{top} & = [2\sin{((e^{1.2t} + 60)t)}],\\
    u_{bottom} & = [2\sin{((e^{1.2t} + 50)t)}].
\end{align*}

\subsubsection*{System 3: Flow Around a Cylinder} For our third system, we consider the 2D flow around a cylinder modeled using the Navier-Stokes equations. The cylinder has a $1$$m$ radius and is centered at $(20,20)$ in a $50 m \times 40 m$ rectangular workspace. For the actual system, $M_{act}$, the cylinder moves vertically along the $y = 20$ axis such that its center moves periodically between $(20,21)$ and $(20,19)$ at a frequency of $0.3927$ $rad/sec$.  The velocity profile at the left boundary is set to be a uniform stream while a zero pressure outflow condition is imposed at the right boundary. The Reynolds number is set to $200$.  In this scenario, the system model or dynamics, $M_{curr}$, is assumed to be that of the stationary cylinder placed in the same uniform free stream flow, at the same location, with the same radius, operating at the same Reynolds number.  We note that the oscillation frequency for the moving cylinder in $M_{act}$ is set to be approximately the vortex shedding frequency of $M_{curr}$.  

%In this case, $M_{curr}$ represents the classic wake behind the cylinder problem, where the cylinder is stationary, while $M_{act}$ represents a case where the cylinder executes a periodic motion perpendicular to the input flow direction (see Fig. ). In this problem, initial conditions, boundary conditions as well as external forcing are considered identical across the two models, and the differences between $M_{curr}$ and $M_{act}$ arise due to differences in the actual physical phenomena.

\subsection{Implementation}
The details of each system's architecture are summarized in Tables \ref{exp1-model}, \ref{exp2-model}, and \ref{exp3-model}. We use Adam \cite{adamoptim}, a powerful and computationally efficient optimization algorithm with the recommended default  parameters to initialize the algorithm. We set the algorithm batch size to 64, and used the Python package Keras \cite{keras} to train the network for a total of 50 epochs. Note that for our dense layers, we chose ReLU as our activation function. The function demonstrated the best performance on our tasks.

\begin{table}[H]
\centering
\resizebox{\columnwidth}{!}{
\begin{tabular}{|l|l|l|}
\hline
\textbf{Layer Kind}                              & \textbf{Activation Function} & \textbf{Number of Nodes} \\ \hline
Input 0: $U_{curr}$, Coordinates                    & N/A                      & N/A                      \\ \hline
Layer 1: TDDL{[}Input 0{]} & ReLU                     & 32                       \\ \hline
Layer 2: LSTM Layer{[}Layer 1{]}                   & Tanh/Sigmoid             & 64                       \\ \hline
Layer 3: LSTM Layer{[}Layer 2{]}                   & Tanh/Sigmoid             & 32                       \\ \hline
Layer 4: LSTM Layer{[}Layer 3{]}                   & Tanh/Sigmoid             & 32                       \\ \hline
Layer 5: Dense Layer{[}Layer 4{]}                  & ReLU                     & 10                       \\ \hline
Layer 6: Dense Layer {[}Layer 5{]}                 & Linear                   & 1                        \\ \hline
\end{tabular}
}
\caption{Neural network parameters for System 1. Note that TDDL stands for Time Distributed Dense Layer, LSTM stands for Long Short-Term Memory, and ReLU stands for Rectified Linear Unit.}
\label{exp1-model}
\end{table}

\begin{table}[H]
\centering
\resizebox{\columnwidth}{!}{
\begin{tabular}{|l|l|l|}
\hline
\textbf{Layer Kind}                              & \textbf{Activation Function} & \textbf{Number of Nodes} \\ \hline
Input 0: $U_{curr}$,$F$, Coordinates                  & N/A                      & N/A                      \\ \hline

Layer 1: TDDL{[}Input 0{]} & ReLU                    & 32                       \\ \hline

Layer 2:TDDL{[}Layer 1{]} & ReLU                     & 64                       \\ \hline
Layer 3: LSTM Layer{[}Layer 2{]}                   & Tanh/Sigmoid             & 64                       \\ \hline
Layer 4: LSTM Layer{[}Layer 3{]}                   & Tanh/Sigmoid             & 32                       \\ \hline
Layer 5: LSTM Layer{[}Layer 4{]}                   & Tanh/Sigmoid             & 32                       \\ \hline
Layer 6: Dense Layer {[}Layer 5{]}                 & ReLU                     & 10                      \\ \hline
Layer 7:Dense Layer {[}Layer 6{]}                 & Linear                   & 2                       \\ \hline
\end{tabular}
}
\caption{Neural network parameters for System 2. Note that TDDL stands for Time Distributed Dense Layer, LSTM stands for Long Short-Term Memory, and ReLU stands for Rectified Linear Unit.}
\label{exp2-model}
\end{table}

\begin{table}[H]
\centering
\resizebox{\columnwidth}{!}{
\begin{tabular}{|l|l|l|}
\hline
\textbf{Layer Kind}                                & \textbf{Activation Function} & \textbf{Number of Nodes} \\ \hline
Input 0: $U_{curr}$, Coordinates, Cylinder Position & N/A                      & N/A                      \\ \hline
Layer 1: TDDL{[}Input 0{]}   & ReLU                     & 32                       \\ \hline
Layer 2: TDDL{[}Layer 1{]}   & ReLU                     & 64                       \\ \hline
Layer 3: LSTM Layer{[}Layer 2{]}                     & Tanh/Sigmoid             & 64                       \\ \hline
Layer 4: LSTM Layer{[}Layer 3{]}                     & Tanh/Sigmoid             & 32                       \\ \hline
Layer 5: LSTM Layer{[}Layer 4{]}                     & Tanh/Sigmoid             & 32                       \\ \hline
Layer 6: Dense Layer {[}Layer 5{]}                   & ReLU                     & 10                       \\ \hline
Layer 7: Dense Layer {[}Layer 6{]}                   & Linear                   & 2                        \\ \hline
\end{tabular}
}
\caption{Neural network parameters for System 3. Note that TDDL stands for Time Distributed Dense Layer, LSTM stands for Long Short-Term Memory, and ReLU stands for Rectified Linear Unit.}
\label{exp3-model}
\end{table}

Given the lightweight nature of our networks and the small size of input data, we trained the networks on a CPU Intel(R) Core(TM) i7-8750H CPU @ 2.20GHz. Tensorflow, the backend of Keras, automatically distributes training on multiple cores.  The average time for completing one epoch for Systems 1, 2, and 3 is 1, 60, and 40 seconds respectively.  The differences in training time between each system is mostly due to the training set size. The marginal difference  between each system architecture does not significantly change the training time.

It is important to note that expanding the neural net input size will impact the computational time. Adding more points to the hypercube will result in $d$ more connections where $d$ is the number of nodes in the Stage 1 first layer. These $d$ new connections represent the new input contribution to each node in the first layer. We can also apply the network on longer data sequences. This would not result in any new connections, but it will result in applying Stages 1 and 2 of the network on the added time steps. Both of these changes, when studied independently, will result in a constant increase in the number of operations for both prediction and training.

There is also an impact on computational time through the addition of more data. In training neural networks, we apply the same vectorized operations, mostly matrix multiplications, on batches of data. The nature of this computational process means that for each new data point, the number of operations for both training and prediction increases by a constant factor.

Finally we note that in solving new problems, we might need to expand the network representational capacity by adding more nodes and layers. The change in the computational cost of the network will heavily depend on the size and complexity of the new network. However, recent advances in GPU development tailored specifically for Deep Learning offers a range of solutions for building optimized and scalable implementations of complicated and heavy architectures.

\subsection{Datasets}
%\ani{@Mann: Please briefly describe how you generated the data sets for systems 1 and 2.  Basically just briefly describe the numerical techniques you used to solve the problems.  I'll put in the description for OpenFoam.}

In this work, we employ numerical solutions to the actual and assumed models, $M_{act}$ and $M_{curr}$, to generate the ground truth, test, and training data sets.  The ground truth and actual system observations, $U_{act}$, are obtained by numerically solving $M_{act}$. %The ground truth for the actual system observations and solution, $U_{act}$, are obtained by numerically solving $M_{act}$.  
Similarly, the values for $U_{curr}$ are obtained by numerically solving $M_{curr}$ for the assumed parameter values, e.g., $\tilde{{\cal N}}$. %vs. ${\cal N}$.  
For System 1, the 1D heat equation given by Eq. \eqref{eqn_sec_prob_1D_heat} was solved using the finite volume based PDE solver in Python (FiPy).  The equation was discretized on a spatial $50 \times 1$ grid over $0.006$ seconds with time step of $0.000012$. The number of time frames is $500$. For System 2, a finite difference scheme was used to solve the Navier Stokes equations given by \eqref{eqn_sec_prob_lid_cav_M_act}-\eqref{eqn_sec_prob_lid_cav_M_curr}.  The equation was discretized on a spatial $30 \times 30$ grid over $2$ seconds with time step of $0.001$. The number of time frames is $2000$. Lastly, numerical solutions for System 3, flow around a cylinder, were obtained using OpenFoam \cite{OpenFoamVortexTutorial}.  The datasets for $U_{act}$ and $U_{curr}$ were obtained on a largely uniform grid consisting of approximately $100 \times 70$ points over $2000$ seconds using OpenFoam's {\tt pimpleFoam} solver with a solution time step of $0.001$ seconds.  However, $U_{act}$ and $U_{curr}$ consist of data obtained at every second over the total $2000$ second simulation run and thus the data consist of $2000$ time frames. %The static cylinder with a Reynolds number of 200 in the presence of turbulence is assumed to be $M_{curr}$ while the oscillating cylinder with the same Reynolds number was set $M_{act}$.

Training and test datasets are comprised of both $U_{curr}$ and $U_{act}$. %In this work, we assume measurement error is negligible and thus our observations are given by $\hat{U}_{act} = U_{act}$. The impact of measurement noise on the performance of our system is a direction for future investigation.  
Assuming $U_{act}$ and $U_{curr}$ consists of $N$ frames ($N_x \times N_y$ grid points in each frame), we partition the data into  four sets: training, validation, local test, and future test. For the training, validation, and local test sets, we consider $K$ consecutive frames. In order to split the data in the $K$ frames between the three sets, we split the set of $N_x \times N_y$ grid points at each frame randomly between training, validation and local test sets. We choose to include $60\%$ of the grid points in the training set, 10\% of the grid points in the validation set, and $30\%$ of the grid points in the local test set. Note that the split is the same for each frame. For the future test set, we consider the remaining $T = N - K$ frames with all the grid points. We use the training set to train the network, the validation set to test the effect of hyperparameter optimization and different network architectures on the network performance, the local test set to measure the chosen network ability to generalize over unseen grid points, and the future test set to measure the network ability to generalize over unseen dynamics, {\it i.e.}, predict future observations. 

For System 1, we include the first 150 frames in the training, validation, and local test sets ($ K=150$) and the last 350 frames ($T=350$) in the future test set. This setup results in a total of 4,144 points for training, 740 points for validation, 2,220 for local test, and 16,800 for future test. For System 2, we use the first 1,000 frames for training ($ K=1,000$), validation and local test and use the last 1,000 ($T=1,000$) frames for the future test. This results in a total of 469,060 points for training, 77,844 points for validation, 235,528 points for local test, and 784,000 points for future test. For System 3, we use 100 frames of data between the frames $500$ and $600$, \ie between $500$ and $600$ seconds, for the training, validation and local test set. We used the 1,400 frames after 600 seconds for the future test. This results in a total of 399,800 points for training, 66,600 for validation, 200,000 points for local test, and 9,786,000 points for future test. 

%\ani{@Maan: Can you update the last sentence to match the first 2?  So the third sentence should be: For System 3, ... }

\subsection{Evaluation Metrics}\label{subsec_eval}
To assess model performance, we use two benchmarks to measure the difference between two sets of $F$ frames: $Set_{1}$ (tested set) and $Set_{2}$ (ground truth). In our evaluation, $Set_{1}$ will either be $U_{curr}$ or $U_{nn}$ and $Set_{2}$ will be $U_{act}$. %\ani{@Maan: Is there a way to define $Set_{1}$ and $Set_{2}$ in terms of $M_{act}$ and $M_{curr}$ or $U_{act}$ and $U_{curr}$?} 

\paragraph{Mean Squared Errors (MSE)} The first benchmark uses the mean squared difference between $Set_{1}$ and $Set_{2}$. For 2D output, we take the average of the two outputs for every point before computing the MSE. We also include the mean magnitude square difference (MMSD) and the mean cosine similarity (MCS) between the two sets in the benchmark for the Lid Cavity and the Flow Around a Cylinder problems in Systems 2 and 3 since both are 2D systems. 

\paragraph{Proper Orthogonal Decomposition} The second benchmark compares the proper orthogonal decomposition (POD) modes that accounts for $99\%$ of the system variation.  Complex nonlinear dynamical systems can exhibit significant spatiotemporal variations, often at differing scales.  To extract the dominant dynamics of these systems, techniques for modal analysis are often used to construct a reduced order representation of the dynamics. POD is a data-driven reduced order modeling strategy that is often used to identify the dominant dynamics of a system purely from observations \cite{Sirovich1987TurbulenceStructures, KirbyBook2000}.

Given $N$ snapshots of the system states which can be obtained either through measurements and/or numerical simulations, let $\bm{x(t)} = [x_1(t),...,x_k(t)]^\top$ denote the set of spatial coordinates in $\mathbb{W}$ at $t = 1,\ldots,N$.  We note that the points in $\bm{x(t)}$ correspond to the grid points in which $u_{act}$ and $u_{curr}$ values are provided at some given time $t$.  Using $\bm{x(t)}$, we can construct a covariance matrix as
\begin{equation}\label{cov_mat}
	\bm{K} =  \frac{1}{m} \sum_{t=1}^{m} \bm{x(t)x(t)}^\top = \frac{1}{m} \bm{XX}^\top,
\end{equation}
where $\bm{X} \in \mathbb{R}^{n \times m}$ with its columns as $\bm{x(t)}$.  To extract the dominant dynamic modes from the data given by $\bm{x(t)} = [x_1(t),...,x_k(t)]^\top$ for $t = 1,...,N$, we obtain the low dimensional basis for the data by solving the symmetric eigenvalue problem
\begin{equation*}
	\bm{K \phi_{i}} =  \lambda_{i} \bm{\phi_{i}},
\end{equation*}
where $\bm{K}$ has $N$ eigenvalues such that $\lambda_{1} \geq \lambda_{2} ... \geq \lambda_{N} \geq 0$ and the eigenvectors $\bm{\phi}$ are pairwise orthonormal.

The original basis is then truncated into a new basis $\bm{\Phi}$ by choosing $k$ eigenvectors that capture the desired fraction, $E$, of the total variance of the system, such that their eigenvalues satisfy
\begin{equation*}
	\frac{\sum_{i=1}^{k} \lambda_i}{\sum_{i=1}^{n} \lambda_i}  \geq E.
\end{equation*}
Each term $\bm{x(t)}$ can be written as
\begin{equation} \label{field}
	\bm{x(t)} = \bm{\Phi c(t)},
\end{equation}
where $\bm{c(t)} = [c_1(t), ..., c_k(t)]^\top$ holds time-dependent coefficients and $\bm{\Phi} \in \mathbb{R}^{n \times k}$ with its columns as $\bm{\phi_1}$,...,$\bm{\phi_k}$. The low-dimensional, orthogonal subspace associated with $\bm{\Phi}$ is an optimal approximation of the data with respect to minimizing least squares error.

To compare the POD modes, we compute the inner product, {\it i.e.} the cosine similarity, between the two sets of principal components obtained for $Set_{1}$ and $Set_{2}$. We call this metric CS-POD, for short. We calculate the statistics of both benchmarks on two cases: Case 1, $Set_1$ is $M_{curr}$ and Case 2, $Set_1$ is $M_{nn}$. Case 1 provides a relative baseline for measuring the performance of the neural net in Case 2. We report the first benchmark statistics over training, local test, and future test sets, and the second benchmark statistics over the entire simulation. 

%based on a PCA analysis. we compute the $R$ principal components of $Set{2}$ (ground truth) that accounts for $99\%$ of the variation. We then consider the $R$ principal components of $Set_{1}$ and compute the Cos Similarity between the two sets of $R$ principal components. 

\section{Results and Discussion}\label{sec_results}
We present and discuss the results of our proposed learning framework for each of the candidate systems. 

\subsubsection*{System 1: 1D Heat Equation}
\begin{figure*}[ht]
    \centering
    \subfloat[]{\includegraphics[width = 0.45\linewidth]{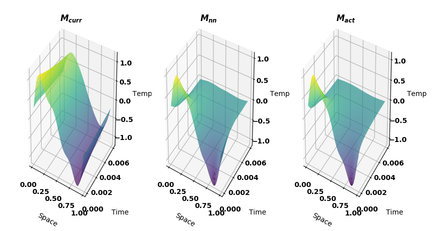}\label{exp1_fig}}
    \subfloat[]{\includegraphics[width = 0.275\linewidth]{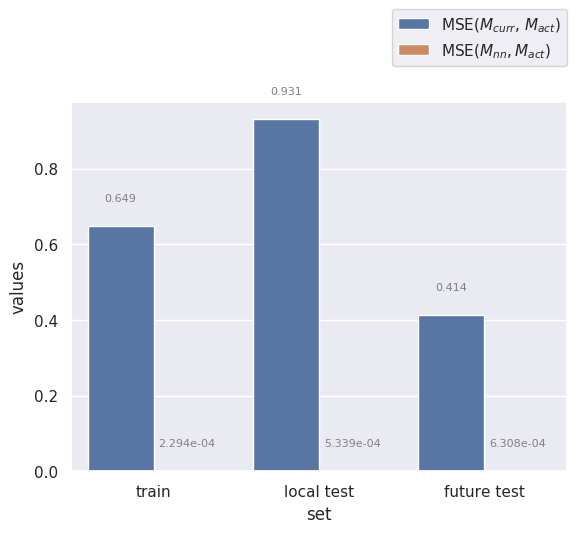}\label{exp1_b1}}
    \subfloat[]{\includegraphics[width = 0.275\linewidth]{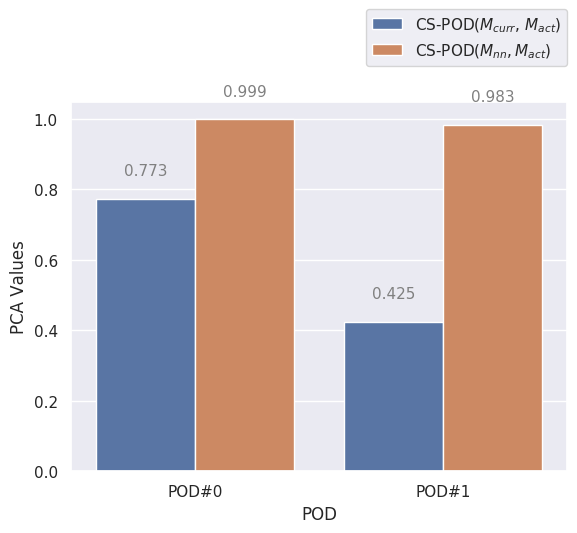}\label{exp1_b2}}
    \caption{(a) System 1: Temperature as a function of the spatial and temporal coordinates for (left) $U_{curr}$, (middle) $U_{nn}$, and (right) $U_{act}$. (b) MSE between $U_{act}$ and $U_{curr}$ (light blue) and $U_{act}$ and $U_{nn}$ (dark blue) for system 1. (c) Cosine similarity between the first principal POD mode of $U_{act}$ and $U_{curr}$ (light blue) and $U_{act}$ and $U_{nn}$ (dark blue) for system 1.}
    \label{fig:system1}
\end{figure*}

 Figure \ref{exp1_fig} shows the temperature data generated by $M_{act}$, $M_{curr}$, and $M_{nn}$ for the entire spatiotemporal domain.  In these simulations, $D_{act}$ and $D_{curr}$ were set to $15$ and $1$ $mm^2/s$ respectively. Qualitatively we see that the network model does an excellent job in resolving the inaccurately modeled dynamics and accurately captures the true dynamics of the system. Figure  \ref{exp1_b1} quantitatively shows the network's ability to generalize over local unseen grid points as well as data in the future set. In fact, one can see that the error between $U_{act}$ and $U_{nn}$ is orders of magnitudes less than that of $U_{act}$ and $U_{curr}$. Moreover, the error bars between $U_{act}$ and $U_{nn}$ are so small that the orange bars are not visible in the graph (the exact values for comparison are denoted in the figure). The quantitative results are further confirmed in Fig. \ref{exp1_b2} which shows the CS-POD for the POD modes. In this problem, the POD decomposition of $U_{nn}$ over the entire simulation yielded two principal modes as did the POD decomposition of $U_{act}$. The agreement between these POD modes is excellent as demonstrated by a CS-POD values that are very close to unity, as seen in  Fig. \ref{exp1_b2}. 

\subsubsection*{System 2: Lid Cavity Problem}
\begin{figure}[H]
    \centering
    \includegraphics[width = 0.9\linewidth]{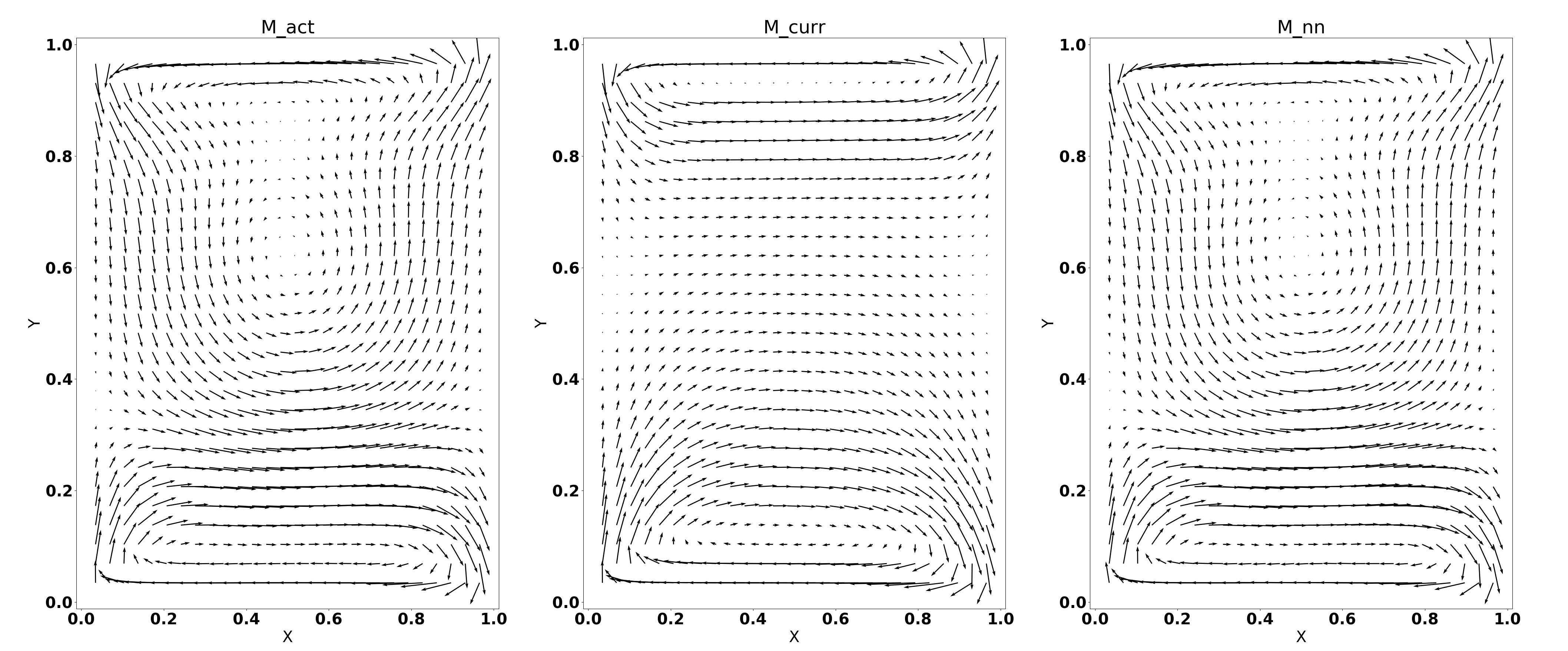}
    \caption{System 2:Vector field at $t=1.147$ seconds for (left) $U_{curr}$, (middle) $U_{nn}$, and (right) $U_{act}$.} 
    \label{exp2_fig1}
\end{figure}

\begin{figure*}
    \centering
    \subfloat[]{\includegraphics[width=0.49\linewidth]{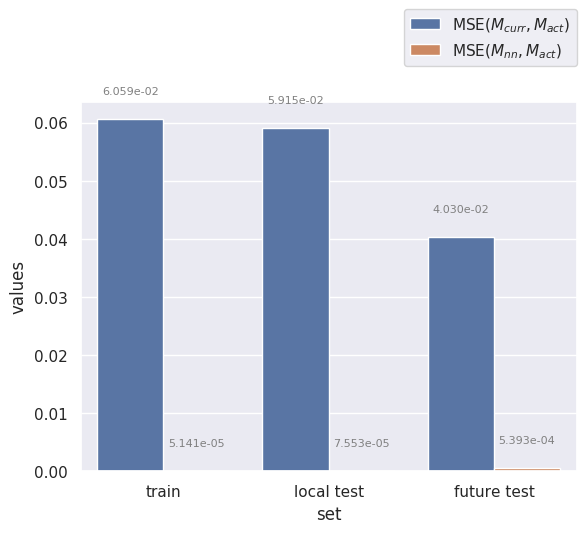}\label{exp2_b1_mse}}
    \subfloat[]{\includegraphics[width=0.49\linewidth]{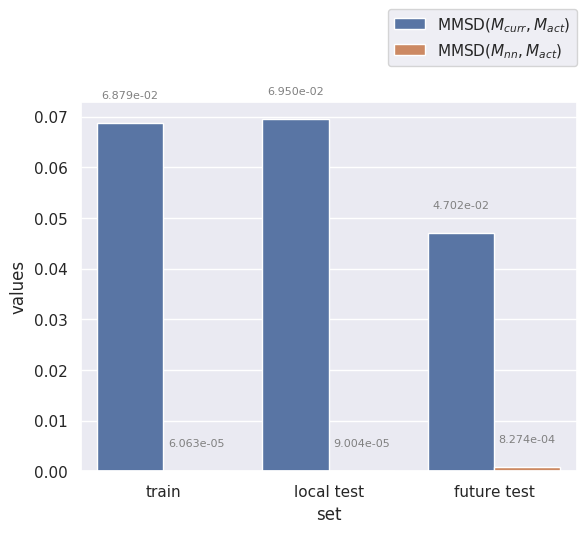}\label{exp2_b1_MMSD}}\\
    \subfloat[]{\includegraphics[width =0.49\linewidth]{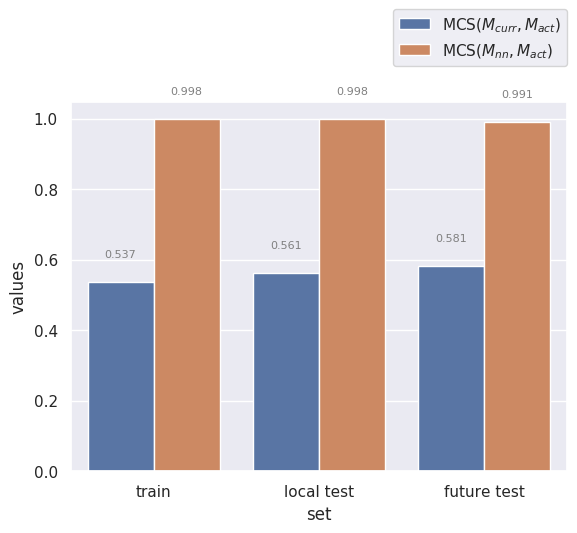}\label{exp2_b1_mcs}}
    \subfloat[]{\includegraphics[width =0.49\linewidth]{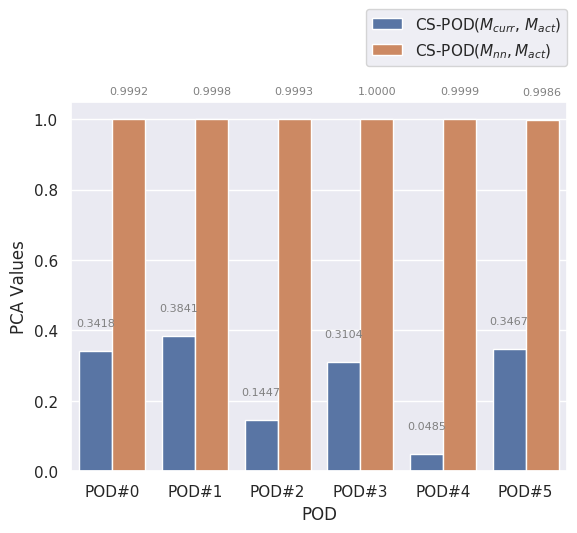}\label{exp2_b2}}
    \caption{(a) MSE between $U_{act}$ and $U_{curr}$ (light blue) and $U_{act}$ and $U_{nn}$ (dark blue) for system 2. (b) MMSD  between $U_{act}$ and $U_{curr}$ (light blue) and $U_{act}$ and $U_{nn}$ (dark blue) for system 2. (c) Mean cosine similarity between $U_{act}$ and $U_{curr}$ and $U_{act}$ and $U_{nn}$ for system 2. (d) Cosine similarity between the first five POD modes of $U_{act}$ and $U_{curr}$ (light blue) and $U_{act}$ and $U_{nn}$ (dark blue) for system 2.}
    \label{fig:system2_compare}
\end{figure*}

  Figure \ref{exp2_fig1} shows a snapshot of the vector field generated by $M_{act}$, $M_{curr}$, and $M_{nn}$ for the entire domain at $t=1.147$ seconds. As with System 1, we see qualitatively that the network model does an excellent job in resolving the inaccurately modeled dynamics and does accurately capture the actual dynamics of the system. Figures \ref{exp2_b1_mse}-\ref{exp2_b1_mcs} quantitatively show the network's ability to combine $U_{curr}$ and observations of $g_1 = F$ to correctly predict $U_{act}$ over local unseen grid points as well as data in the future set.  Figures \ref{exp2_b1_mse} and \ref{exp2_b1_MMSD} show respectively that the MSE and MMSD between $U_{act}$ and $U_{nn}$ is orders of magnitudes less than that of $U_{act}$ and $U_{curr}$. The quantitative results are further
confirmed in Figs. \ref{exp2_b1_mcs} which shows that the mean cosine similarity between $U_{act}$ and $U_{nn}$ are close to unity thus demonstrating that $U_{nn}$ is resolving the actual dynamics to a far greater degree than is $U_{curr}$. Similarly, Fig. \ref{exp2_b2} shows the CS-POD for the first six principal POD modes over the entire simulation. The CS-POD values again demonstrate the network's ability to resolve the actual system's dynamics. In short, the high degree of accuracy  shows that our network is capable of correctly predicting observations both in previously unseen regions in the workspace as well as in future time steps.

 %observation is further confirmed from results in Fig. \ref{exp2_b2}, where $U_{nn}$ 5 POD modes over the future test set matched those of $U_{act}$ to a great accuracy. 

%\ani{@Maan: Can you make the png wider in Fig. \ref{exp2_fig1}?  Sometimes it helps to manually cut out the white space around the figures if there is any.  Also, can you show from left to right $U_{act}$, $U_{curr}$, and $U_{nn}$?}

%\begin{figure*}
%    \centering
%    \begin{minipage}{0.5\textwidth}
%        \centering
%       \includegraphics[width=0.9\textwidth]{figs/exp2_b1_mse.png}
%        \caption{Experiment 2, MSE  between $U_{act}$, $U_{curr}$ and $U_{act}$, $U_{nn}$}
%        \label{exp2_b1_mse}
%    \end{minipage}\hfill
%    \begin{minipage}{0.5\textwidth}
%        \centering
%        \includegraphics[width=0.9\textwidth]{figs/exp2_b1_mmsd.png}
%        \caption{Experiment 2, MMSD between $U_{act}$, $U_{curr}$ and $U_{act}$, $U_{nn}$}
%        \label{exp2_b1_MMSD}
%    \end{minipage}
%    \begin{minipage}{0.5\textwidth}
%        \centering
%        \includegraphics[width =0.9\textwidth]{figs/exp2_b1_mcs.png}
%        \caption{Experiment 2, MCS between $U_{act}$, $U_{curr}$ and $U_{act}$, $U_{nn}$}
%        \label{exp2_b1_mcs}
%    \end{minipage}\hfill
%    \begin{minipage}{0.5\textwidth}
%        \centering
%        \includegraphics[width =0.9\textwidth]{figs/exp2_b2.png}
%        \caption{Experiment 2: Benchmark 2 Cosine Similarity between the POD modes of $U_{act}$, $U_{curr}$, and $U_{act}$, $U_{nn}$}
%        \label{exp2_b2}
%     \end{minipage}
%\end{figure*}

\begin{figure*}
    \centering
    \includegraphics[width = 0.9\linewidth]{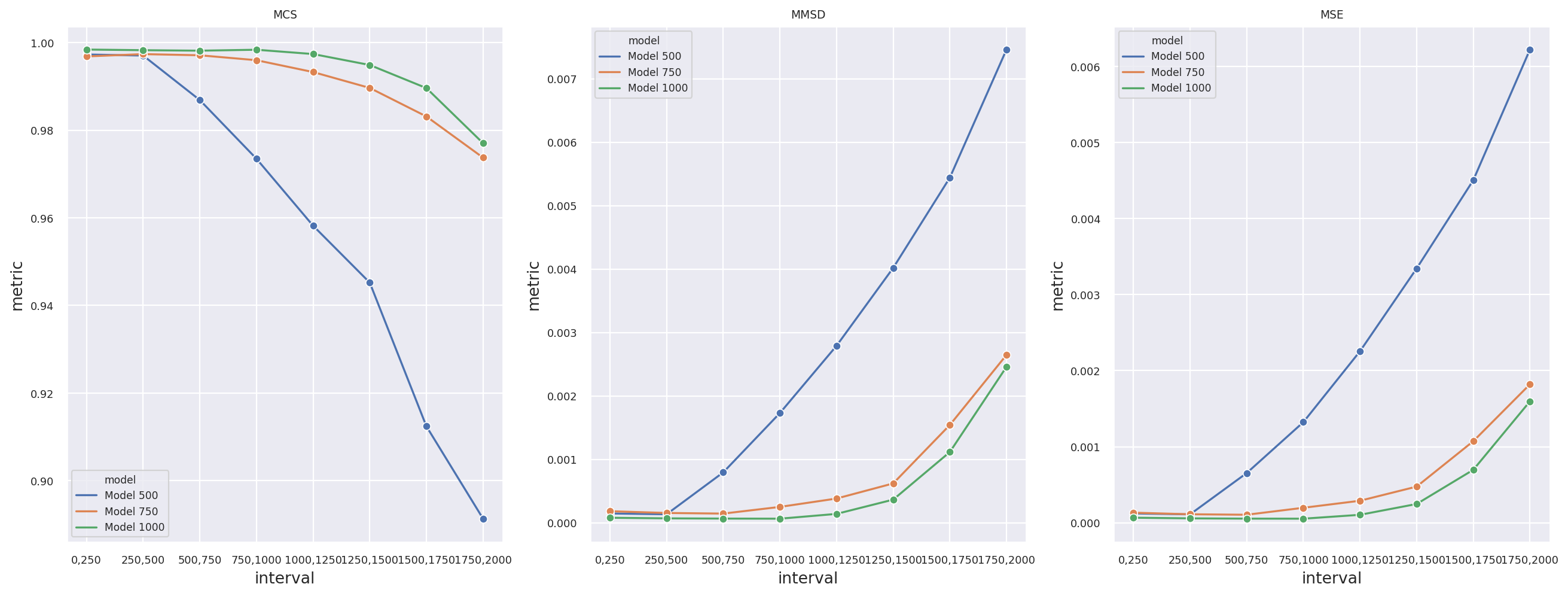}
    \caption{Comparison between the predictive power of the $M_{nn}$ trained using the first $500$, $750$, and $1000$ frames. The $x$ axis denotes the time interval in increments of $250$.  The metric for each interval represents the metric value computed for the frames in that interval.} 
    \label{pre_power}
\end{figure*}

%\noindent\changeto{\textbf{Prediction power analysis:}}
It is important to note that the periods of the body force and the moving upper and lower boundaries in System 2 are not constant.  In fact, they change exponentially as a function of time.  Ideally, the trained network should capture this exponential change in the periods and be able to accurately predict future values outside of the training frames. In reality though, the prediction accuracy would degrade the farther out the prediction times are from the training times. To quantify this behaviour, three training regimes were considered with different training set lengths. The training sets for the three regimes contained the first 500 frames, first 750 frames, and the first 1000 frames of the data set respectively. We evaluated the system's predictive power using intervals of 250 future output frames and the results are shown in Fig. \ref{pre_power}. The metric (MSE, MCS and MMSD) for each interval is computed across all 250 frames in that interval. As expected, the prediction accuracy degrades the further out the prediction time is from the training set. For this particular case, the network is able to predict approximately one training period into the future, with a fair degree of accuracy. 

\subsubsection*{System 3: Flow Around the Cylinder} 
\begin{figure}
    \centering
    \includegraphics[width = 0.7\linewidth]{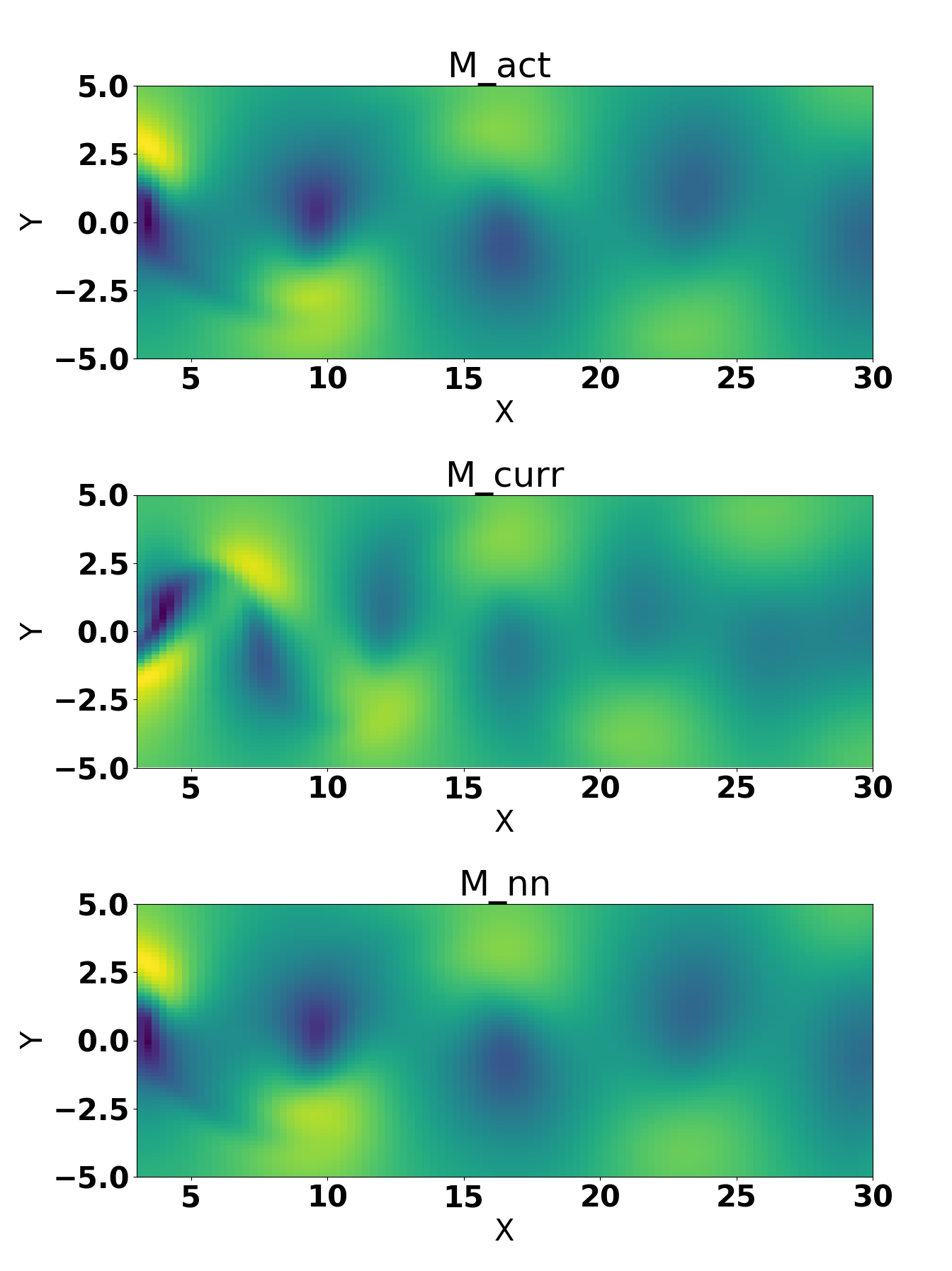}
    \caption{System 3: Magnitude of the velocity profile given by (left) $U_{curr}$, (middle) $U_{nn}$, and (right) $U_{act}$ at time $t= 1390s$.} 
    \label{exp3_fig1}
\end{figure}

\begin{figure*}
    \centering
    \subfloat[]{\includegraphics[width =0.49\linewidth]{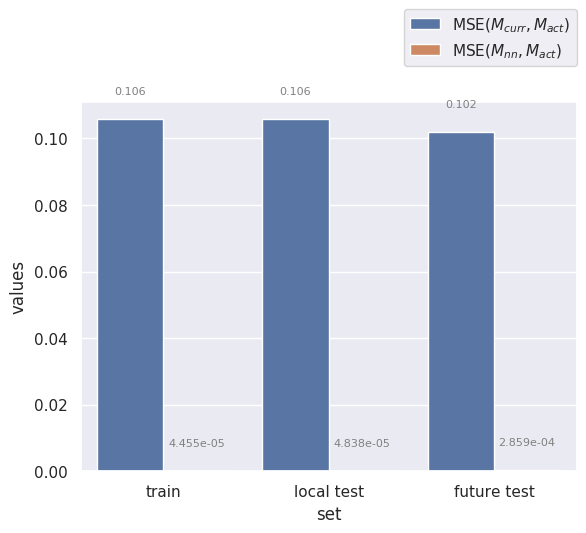}\label{exp3_b1_mse}}
    \subfloat[]{\includegraphics[width =0.49\linewidth]{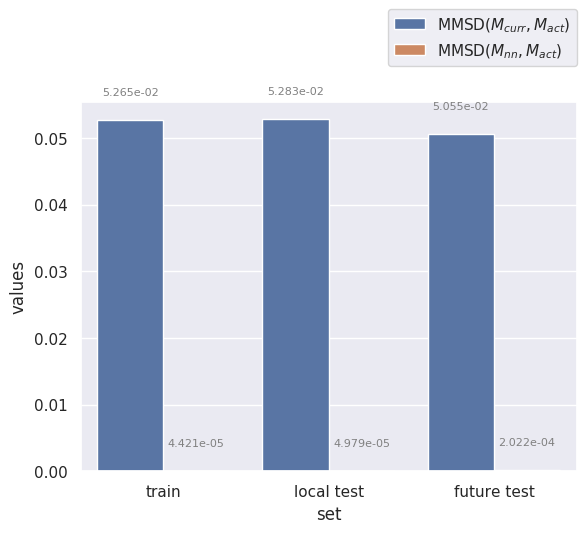}\label{exp3_b1_mmsd}}\\
    \subfloat[]{\includegraphics[width =0.49\linewidth]{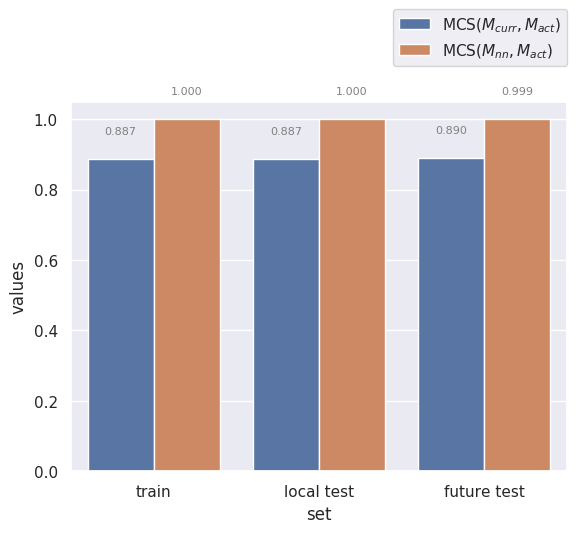}\label{exp3_b1_mcs}}
    \subfloat[]{\includegraphics[width =0.49\linewidth]{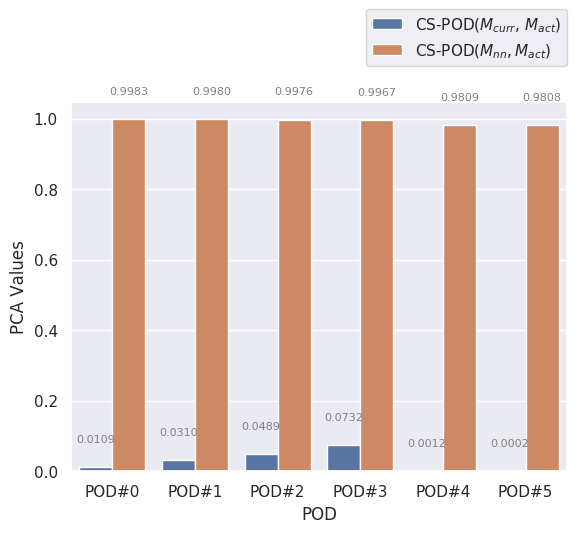}\label{exp3_b2}}
    \caption{(a) MSE between $U_{act}$ and $U_{curr}$ (light blue) and $U_{act}$ and $U_{nn}$ (dark blue) for system 3. (b) MMSD  between $U_{act}$ and $U_{curr}$ (light blue) and $U_{act}$ and $U_{nn}$ (dark blue) for system 3. (c) Mean cosine similarity between $U_{act}$ and $U_{curr}$ and $U_{act}$ and $U_{nn}$ for system 3. (d)Cosine similarity between the first six POD modes of $U_{act}$ and $U_{curr}$ (light blue) and $U_{act}$ and $U_{nn}$ (dark blue) for system 3.}
    \label{fig:system3_compare}
\end{figure*}

 Figure \ref{exp3_fig1} shows a snapshot of the magnitude of the velocity field at $t=1390$ seconds generated by $M_{act}$, $M_{curr}$, and $M_{nn}$. In these results, the Reynolds numbers for both $M_{act}$ and $M_{curr}$ were set to $200$.  As with the previous two systems, we see qualitatively that the network model does an excellent job in resolving the inaccurately modeled dynamics and does accurately capture the actual dynamics of the system. In particular, note that the network model $M_{nn}$ is accurately capturing the vortex shedding frequency while the $M_{curr}$ vortices are out of phase with the actual vortex shedding pattern.  As in the previous systems, Figs. \ref{exp3_b1_mse}-\ref{exp3_b1_mcs} quantitatively show the network's ability to combine values of $U_{curr}$ as well as $g_1 = O(t)$ (where $O$ indicates the position of the cylinder at time $t$) to correctly predict $U_{act}$ over local unseen grid points as well as data in the future test set. Figures \ref{exp3_b1_mse} and \ref{exp3_b1_mmsd} show respectively that the MSE and MMSD between $U_{act}$ and $U_{nn}$ is orders of magnitudes less than that of $U_{act}$ and $U_{curr}$. The quantitative results are further confirmed in Figs. \ref{exp3_b1_mcs} which shows that the mean cosine similarity between $U_{act}$ and $U_{nn}$ are close to unity thus demonstrating that $U_{nn}$ is resolving the actual dynamics to a far greater degree than is $U_{curr}$. Similarly, Fig. \ref{exp3_b2} shows the CS-POD for the first six principal POD modes over the  entire simulation. The CS-POD values again demonstrate the network's ability to resolve the actual system's dynamics. In short, the high degree of accuracy shows that our network is capable of correctly predicting observations both in previously unseen regions in the workspace as well as in future time steps for systems exhibiting more complex dynamics.

%\ani{@Maan: Can you stack the individual images in the png wider in Fig. \ref{exp3_fig1} vertically instead of horizontally?  Vertical stacking would make them bigger.  Also, can you show from top to bottom $U_{act}$, $U_{curr}$, and $U_{nn}$?}

%\noindent\changeto{\textbf{Periodicity analysis: }}

To evaluate the predictive performance of $M_{nn}$, we focus on the network's ability to identify the periodicity of the oscillations. Since System 3 is periodic, once the network learns the true periodicity of the dynamics, it has effectively learned the true dynamics of the system for all future times. To quantify the difference in periodicity between the model output and the ground truth, the following analysis was performed. For each point in the local test set, $\tau_i$, we consider its time series from frame $600$, the last frame in the training set, to frame $2000$ in both $U_{nn}$ and $U_{act}$. We denote these as $U_{nn}(\tau_i, 600-2000)$ and $U_{act}(\tau_i, 600-2000)$ respectively. We start by computing the frequency spectrums of $U_{nn}( \tau_i,600-2000)$ and $U_{act}(\tau_i,600-200)$ using the Fast Fourier Transform (FFT) which we denote as $FFT_{nn}$ and $FFT_{act}$.  Consider the percentage mean absolute difference between the frequencies that corresponds to the energy peaks between $FFT_{nn}$ and $FFT_{act}$ which we denote by $\%\Delta(FFT_{U_{nn}}, FFT_{U_{act}})$. The mean of $\%\Delta(FFT_{U_{nn}}, FFT_{U_{act}})$ is then computed for every grid point in the local test set which resulted in a value of $0.0239$. This analysis indicates that the neural network output not only accurately captures the periodicity of the underlying phenomena but it is able to correctly identify the global features of the dynamics. In short, once the network captures the periodicity, it can then predict the system’s behavior at any time in the future.

\section{Conclusion and Future Work}\label{sec_conclusion}

We have proposed a data-driven modeling strategy based on a neural network machine learning framework that enables one to overcome improperly or inadequately modeled dynamics for systems that exhibit complex spatiotemporal behavior.  Given a system model that does not accurately capture the true dynamics, our machine learning strategy uses data generated from the improper system model combined with observational data from the actual system to create a neural network model. As we have shown with three complex dynamical systems, the network model that is created is capable of accurately resolving the incomplete or inaccurate dynamics to generate solutions that compare very favorably with the actual dynamics, both in previously unobserved regions as well as for future states. 

Our approach leverages state-of-the-art machine learning frameworks and existing, but limited, knowledge of the physical constraints that drives a process.  The result is an equation-free representation of the system dynamics that encodes a baseline understanding of the underlying physics that drives the process. Since our output is a neural network representation of the system model, the output of our network consists of a set of pointwise inferences and thus is equation-free.  Nevertheless, the output can be fed into existing data-driven model discovery techniques to obtain closed-form equation representations of the dynamical system \cite{Brunton16,lusch2017deep}.

In the future, we plan to perform a detailed analysis on our learning framework performance for different error bounds to better understand acceptable deviations from the true model. Associated with this is the effect of noise, and to this end we plan to investigate how measurement uncertainty in $\hat{U}_{act}$ impacts the performance of $M_{nn}$. Since real-world systems are inherently noisy, we must be able to incorporate noisy observational data while still accurately capturing the system's dynamics. As such, it is important to be able to deal with situations where every observation is subject to a noise that is non-negligible or with situations where one has very noisy outlier observations.  While the impact of noise on a network's performance is well documented and studied in the computer vision literature \cite{ref:Plotz2017}, its impacts on networks modeling more complex phenomena is less well understood. A complete analysis of the effect of noise includes consideration of both additive and multiplicative noise, and involves analyzing simulated systems where deterministic and stochastic elements can be tightly controlled to establish ground truth for comparisons.

By developing methods that can deal with negligible and non-negligible noise, we will enable the study of complex and high-dimensional systems including those found in fluid dynamics and in particular geophysical fluid dynamics.  Fluid flows are complex and exhibit multi-scale phenomena whose dynamics are not at all well-understood.  Even the underlying physical mechanisms for flows are not fully understood. In the future, we plan to use the framework developed in this article to make predictions and estimations. For example, in a geophysical flow, information such as wind forcing or data from depth, may not be included in the models. Even with noisy and sparse observations, we would like to investigate if our framework can be used to accurately resolve the inadequately modeled dynamics. 
 
 \clearpage

\bibliographystyle{plainnat}
\bibliography{bibliography/cas-refs,bibliography/gom_prop,bibliography/denoise, bibliography/learning}

\clearpage

\iffalse
\bio{}
Author biography without author photo.
Author biography. Author biography. Author biography.
Author biography. Author biography. Author biography.
Author biography. Author biography. Author biography.
Author biography. Author biography. Author biography.
Author biography. Author biography. Author biography.
Author biography. Author biography. Author biography.
Author biography. Author biography. Author biography.
Author biography. Author biography. Author biography.
Author biography. Author biography. Author biography.
\endbio

\bio{figs/Dhanushka.jpg}
Author biography with author photo.
Author biography. Author biography. Author biography.
Author biography. Author biography. Author biography.
Author biography. Author biography. Author biography.
Author biography. Author biography. Author biography.
Author biography. Author biography. Author biography.
Author biography. Author biography. Author biography.
Author biography. Author biography. Author biography.
Author biography. Author biography. Author biography.
Author biography. Author biography. Author biography.
\endbio

\bio{figs/pic1}
Author biography with author photo.
Author biography. Author biography. Author biography.
Author biography. Author biography. Author biography.
Author biography. Author biography. Author biography.
Author biography. Author biography. Author biography.
\endbio

\fi

\end{document}